\documentclass[conference]{IEEEtran}
\usepackage{fancyhdr}

\IEEEoverridecommandlockouts
\usepackage{cite}
\usepackage{amsmath,amssymb,amsfonts}
\usepackage{algorithmic}
\usepackage{graphicx}
\usepackage{textcomp}
\usepackage{xcolor}
\usepackage{bm}
\def\BibTeX{{\rm B\kern-.05em{\sc i\kern-.025em b}\kern-.08em
    T\kern-.1667em\lower.7ex\hbox{E}\kern-.125emX}}
\begin{document}

\title{Adaptive Neuron-wise Discriminant Criterion and Adaptive Center Loss at Hidden Layer\\ for Deep Convolutional Neural Network}

\author{\IEEEauthorblockN{Motoshi Abe}
\IEEEauthorblockA{\textit{School of Engineering} \\
\textit{Hiroshima University}\\
Higashi-hiroshima, Japan \\
i13abemotoshi@gmail.com
}
\and
\IEEEauthorblockN{Junichi Miyao}
\IEEEauthorblockA{\textit{Department of Information Engineering} \\
\textit{Hiroshima University}\\
Higashi-hiroshima, Japan \\
miyao@hiroshima-u.ac.jp
}
\and
\IEEEauthorblockN{Takio Kurita}
\IEEEauthorblockA{\textit{Department of Information Engineering} \\
\textit{Hiroshima University}\\
Higashi-hiroshima, Japan \\
tkurita@hiroshima-u.ac.jp}
}

\maketitle

\begin{abstract}
A deep convolutional neural network (CNN) has been widely used in image classification and gives better classification accuracy than the other techniques. 
The softmax cross-entropy loss function is often used for classification tasks.
There are some works to introduce the additional terms in the objective function for training to make the features of the output layer more discriminative.
The neuron-wise discriminant criterion makes the input feature of each neuron in the output layer discriminative by introducing the discriminant criterion to each of the features.
Similarly, the center loss was introduced to the features before the softmax activation function for face recognition to make the deep features discriminative.
The ReLU function is often used for the network as an active function in the hidden layers of the CNN.
However, it is observed that the deep features trained by using the ReLU function are not discriminative enough and show elongated shapes.
In this paper, we propose to use the neuron-wise discriminant criterion at the output layer and the center-loss at the hidden layer.
Also, we introduce the online computation of the means of each class with the exponential forgetting.
We named them adaptive neuron-wise discriminant criterion and adaptive center loss, respectively.
The effectiveness of the integration of the adaptive neuron-wise discriminant criterion and the adaptive center loss is shown by the experiments with MNSIT, FashionMNIST, CIFAR10, CIFAR100, and STL10.

\end{abstract}

\begin{IEEEkeywords}
Convolutional neural network, discriminative feature, center loss, adaptive center loss, neuron-wise neuron-wise discriminant criterion, adaptive neuron-wise neuron-wise discriminant criterion
\end{IEEEkeywords}

\section{Introduction}

Deep convolutional neural network (CNN) have achieved great success for classification problems to improve the state of the art such as object detection and classification \cite{t2,t3,t4,t5,t6}, scene recognition \cite{t7,t8}, action recognition \cite{t9,t10,t11} and so on. 
Usually, the features are extracted by several layers with the convolution filters, and they are used for classification at the output layer.

In the last layer of the network, the softmax function is frequently used for classification tasks because the output of the softmax function can be considered as an approximation of the posterior probability of each class and the decision based on the maximum posterior probability gives the best classification performance in terms of the smallest classification errors from Bayesian decision theory.
Usually, the softmax cross-entropy loss is used as the loss function for the training of the parameters in the deep neural network.
This is equivalent to maximize the likelihood of the estimation of the posterior probabilities for the training samples.

For recognition or classification tasks, the features extracted by the trained CNN need to be not only separable but also discriminative.
Several loss functions have been introduced such as \cite{t12,t1,t13,t15,t17} and so on to make the features discriminative.

Each neuron in the final layer of the CNN with softmax activation function can be thought of as classifying between the class in charge of that neuron and the other classes.
If we consider the role of each neuron to be two-class classification, the neuron is estimating the posterior probability of that class by using a logistic function.
Since the logistic function is obtained as the posterior probability for the case where the probability density functions of two classes are Gaussian with the same variance, it is expected that the probability distributions of the input of the neuron for both the class in charge of that neuron and the other classes become close to the Gaussian distribution.
As pointed out in our previous work \cite{t12}, we can confirm that this phenomenon from the histograms in \cite{t12}.
Ide et al. \cite{t12} proposed to use the neuron-wise discriminant criterion to make these distributions are more separable by using the neuron-wise discriminant criterion for the input of each neuron in the output layer.

Similarly, the center loss \cite{t1} is introduced to the features before the softmax function in the output layer.
This loss is defined by using the distance between the feature vectors and the mean vector of each class.
The minimization of the center loss makes the variances of the feature vectors within each class a minimum.
It is expected that this loss can make the features more discriminative.

Several nonlinear activation functions are introduced at each neuron in the hidden layers or the final output layer of the deep neural network to make the mapping realized by the trained network nonlinear.
There are many activation functions such as binary step function, sigmoid function, ReLU function \cite{t16}, softmax function, and so on.
In the 1990s, the sigmoid function had been frequently used as an activation function at the neurons in the hidden layers of the neural network.
The gradients for the loss function decrease exponentially with the number of layers in gradient-based learning algorithms, such as error-backpropagation, which is known as a gradient vanishing problem.
For deep neural networks, this problem becomes more crucial  because the update of the weights at the front layers in the network becomes very slow.
To prevent the gradient vanishing problem, ReLU function \cite{t16} has been often used as the activation function at the neurons in the hidden layers of the deep neural networks.
But as the side effect of this merit, the output values of the neuron becomes unbounded and may have large values.
It is observed that the input features before the ReLU function in the hidden layers have elongated shapes as shown in Fig. \ref{plot} (a).

In this paper, we propose to use the neuron-wise discriminant criterion at each neuron of the output layer.
Also, we introduce the center loss at the input features of the ReLU function in the hidden layer to prevent the elongated shapes of the hidden features.
Since the neuron-wise discriminant criterion and the center loss work at different locations in the network, it is expected that this combination of two criteria can accelerate and improve the classification accuracy.

Moreover, we introduce the online computation of statistics of the learned features by using the exponential forgetting as the weights of the past samples.
The exponential forgetting makes the weights for past samples smaller with a single parameter for a forgetting factor.
By this online computation, it is possible to efficiently estimate the mean values of two classes for the inputs of each neuron in the outputs layer and the mean feature vectors of each class at the input features of the ReLU function in the hidden layer.
We call each of them the adaptive neuron-wise discriminant criterion and the adaptive center loss, respectively.

The effectiveness of the adaptive neuron-wise discriminant criterion and the adaptive center loss is confirmed by the experiments with FashionMNIST, CIFAR10, CIFAR100, and STL10.

\section{Related Works}

\subsection{Convolutional Neural Networks (CNN)}

Deep convolutional neural networks (CNN) have achieved great success in image classification problems. 
CNN consists of several convolution layers and fully-connected layers.
The computation in the convolution layer is the filtering with the trainable weights. 
The fully-connected layers integrate the features extracted by the convolution layers for classification.
Usually nonlinear activation function is introduced at each neuron in the hidden layers.

In the deep neural networks, it is known that the gradients of the loss function exponentially decrease with the number of layers.
This phenomenon is known as a gradient vanishing problem and is troublesome for the gradient-based learning algorithm, such as the error backpropagation.
To prevent the gradient vanishing problem in the deep neural networks,
ReLU function has been frequently used as the standard activation function of the neurons in the hidden layers.
The ReLU function $\phi(\eta)$ is defined by
\begin{align}
    \label{relu}
    \phi (\eta) = \max(0,\eta) ,
\end{align}
where $\eta$ denotes the input of the ReLU function. 

The softmax function is often used at the output layer for classification and the softmax cross-entropy loss has been widely used as the objective function for training the parameters of the network.

Let $\{ (I_i,\bm{t}_i)|i=1\ldots N\}$ be a set of training samples, where $I_i$ is the $i$-th image and $\bm{t}_i$ is the class label vector of the $i$-th image $I_i$. 
We assume that the class label vector $\bm{t}_i = \begin{bmatrix} t_{i1} & t_{i2} & \cdots & t_{iK} \end{bmatrix}^T$ is represented as one hot vector.
Then the output of the $k$-th neurons in the output layer is defined by using softmax function as
\begin{align}
    \label{softmax}
    y_{ik} = \frac{\exp(z_{ik})}{\sum_{j=1}^K \exp(z_{ij})},
\end{align}
where $\bm{z}_i = \begin{bmatrix} z_{i1} & z_{i2} & \cdots & z_{iK} \end{bmatrix}^T \in \mathbb{R}^K$ and the $k$-th element of the vector $\bm{z}_i$, namely $z_{ik}=z_k(I_i) \in \mathbb{R}$, denotes the input of $k$-th neuron at the output layer for the input image $I_i$.

The softmax cross-entropy loss for the training samples $\{ (I_i,\bm{t}_i)|i=1\ldots N\}$ is defined as
\begin{align}
    \label{softmaxloss}
    \mathcal{L}_S= - \sum_{i=1}^{N} \sum_{k=1}^K t_{ik} \log y_{ik}, 
\end{align}
where $K$, and $N$ are the number of classes, and the number of training samples.

\subsection{Neuron-wise Discriminant Criterion}

We can observe that probability distributions for each class of the input of the neuron at the output layer become close to the Gaussian distribution.
Since we can consider that the neurons of the output layer is doing the binary classification between the target class and the other classes, we can accelerate the discrimination by introducing the neuron-wise discriminant criterion for this binary classification \cite{t12}.

The neuron-wise discriminant criterion is a measure of discrimination between the distributions of each class and is defined as
\begin{align}
    \label{discriminantcriterion}
    \mathcal{L}_D = \sum_{k=1}^{K}\frac{\sigma_{W_k}^2}{\sigma_{T_k}^2} ,
\end{align}
where
\begin{align}
    \label{sigmaw}
    \sigma_{W_k}^2 &= \frac{1}{N}\sum_{i}^{N}\{ t_{ik}(z_{ik}-\mu_k)^2 + (1-t_{ik})(z_{ik}-\hat{\mu}_k)^2\} \\
    \label{sigmat}
    \sigma_{T_k}^2 &= \frac{1}{N}\sum_{i}^{N}(z_{ik}-\mu_{Tk})^2.
\end{align}
The means of the target class and the non-target class of the inputs of $k$-th neuron are denoted by $\mu_{k}$ and $\hat{\mu}_k$ respectively.
The total mean of the inputs of the $k$-th neuron is denoted as $\mu_{Tk}$.
Similarly the number of samples of the $k$-th class is denoted as $N_k=\sum_{i}^{N}t_{ik}$.
Then the number of samples of the non-traget class is given by $\hat{N}_k=\sum_{i}^{N}(1-t_{ik}) = 1 - N_k$.
Then the means are given as
\begin{align}
\mu_k&=\frac{1}{N_k}\sum_{i}^{N}t_{ik}z_{ik}, \ \ \ 
\hat{\mu}_k=\frac{1}{\hat{N}_k}\sum_{i}^{N}(1-t_{ik})z_{ik},\\
\mu_{Tk}&=\frac{1}{N}\sum_{i}^{N}z_{ik}.
\end{align}

The objective function for training the parameters of the network is defined by combining the softmax cross-entropy loss and the neuron-wise discriminant criterion as
\begin{align}
    \label{softdisc}
    \mathcal{L} = \mathcal{L}_S + \lambda \mathcal{L}_D
\end{align}
where $\lambda$ is a hyper parameter to balance the softmax cross-entropy loss and the neuron-wise discriminant criterion.

The neuron-wise discriminant criterion can accelerate discrimination between two classes, but it is necessary to keep all the input values of each neuron to calculate the discriminant criterion.
In this paper, we apply online computation with the exponential forgetting to compute the statistics such as $\sigma_{W_k}$ and $\sigma_{T_k}$.
We call this method the adaptive neuron-wise discriminant criterion, and the details are explained in the next section.

\subsection{Center Loss at Output Layer}

It is known that trained CNN with the softmax cross-entropy loss and the ReLU function leads to better accuracy for image classification and the other classification tasks.
The center loss \cite{t1} was introduced to improve the recognition accuracy for the face recognition tasks further at the output layer similar to the neuron-wise discriminant criterion.

The center loss minimizes the distance between the extracted feature vectors and the mean vector of each class and is defined as
\begin{align}
    \label{centerloss}
    \mathcal{L}_C = \frac{1}{N}\sum_{i=1}^{N}||\bm{z}_i - \bm{c}_{k}||_2^2,
\end{align}
where $\bm{z}_i$ is the feature vector for the $i$-th training image $I_i$ at the output layer, and $\bm{c}_{k}$ denotes the mean vector of the feature vectors for $k$-th class.

The authors proposed the method to update the mean vector using the samples in the mini-batch.
Let $B$ be a set of indexes of the samples in the mini-batch.
Then the update rule of the mean vector $\bm{c}_k$ of the $k$-the class is defined as
\begin{align}
    \label{pc2}
    \bm{c}_k &\leftarrow \bm{c}_k - \beta \Delta \bm{c}_k,
\end{align}
where $\beta \in [0,1]$ is a hyper parameter and
\begin{align}
    \label{pc1}
    \Delta \bm{c}_k &= \frac{\displaystyle \sum_{i\in B}t_{ik}(\bm{c}_k - \bm{z}_i)}{1 + \displaystyle \sum_{i\in B}t_{ik}}.
\end{align}

The objective function for training is defined by the combination of the standard softmax cross-entropy loss and the center loss as
\begin{align}
    \label{softcen}
    \mathcal{L} = \mathcal{L}_S + \lambda \mathcal{L}_C,
\end{align}
where $\lambda$ is a hyper parameter to balance the softmax cross-entropy loss and the center loss.

The center loss can make the feature vectors at the output layer more discriminative. 
In this paper, we apply the center loss at a hidden layer instead of the output layer.
Also, we introduce online computation with the exponential forgetting to compute the mean vectors of each class.
We call this method the adaptive center loss, and the details are explained in the next section.

\section{Adaptive Neuron-wise Discriminant Criterion and Adaptive Center Loss}

\subsection{Basic Idea}

\begin{figure}[tb]
    \centering
    \includegraphics[width=\linewidth]{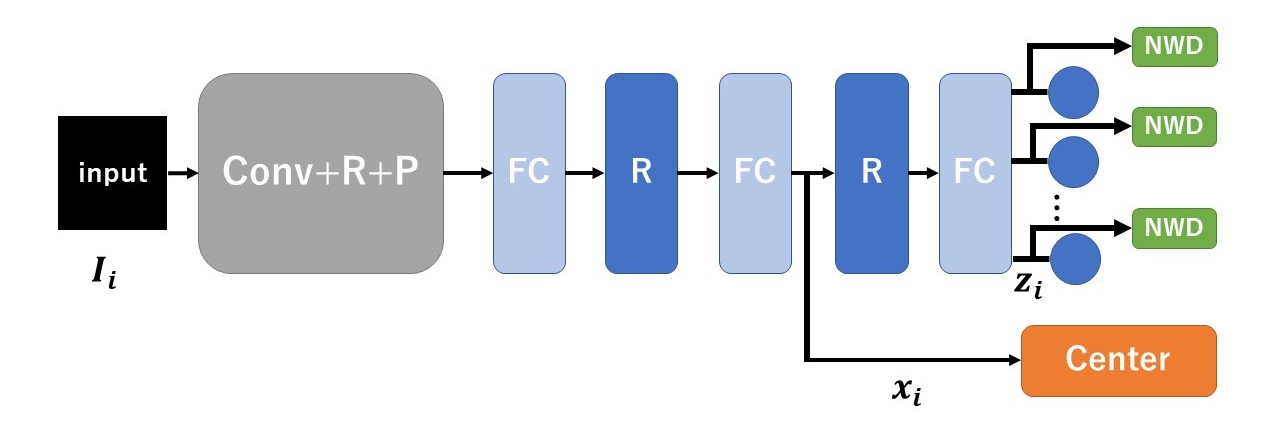}
    \caption{This figure shows the locations of the features of the neuron-wise discriminant criterion and the center loss at the hidden layer on CNN. 
In this figure, Conv+R+P denotes several convolution layers with ReLU activation function and pooling.
 R, P, and FC denote a layer of neurons with the ReLU activation function,  a pooling layer, and a fully connected layer. 
Center and NWD denote the center loss and the neuron-wise discriminant criterion.}
    \label{CNN}
\end{figure}

Fig. \ref{plot} (a) shows the 2-dimensional extracted features before the ReLU function of the hidden layer.
We can observe the elongated shapes of the clusters of each class.
It is expected that this phenomenon can be reduced by introducing the center loss at the hidden layer.
Thus, we proposed to use the center loss at the hidden layer instead of the output layer and to combine it with the neuron-wise discriminant criterion at the output layer.
The locations of the center loss and the neuron-wise discriminant criteria on CNN are shown in Fig. \ref{CNN}.

Also, we introduce the online computation of the means and the variances with the exponential forgetting to calculate the neuron-wise discriminant criterion and the center loss.
We named them the adaptive neuron-wise discriminant criterion and the adaptive center loss, respectively.
The adaptive neuron-wise discriminant criterion can make the input features of the neurons at the output layer discriminative.
The adaptive center loss can also make the extracted feature vectors before the ReLU function at the hidden layer compact and discriminative.

Thus the objective function for the training of the proposed method is defined as
\begin{align}
    \label{adcenaddisc}
    \mathcal{L} = \mathcal{L}_S + \lambda_1 \mathcal{L}_{AD} + \lambda_2 \mathcal{L}_{AC},
\end{align}
where $\lambda_1$, and $\lambda_2$ are the hyper parameters to balance the softmax cross-entropy loss $\mathcal{L}_S$, the adaptive neuron-wise discriminant criterion $\mathcal{L}_{AD}$, and the adaptive center loss $\mathcal{L}_{AC}$, respectively.

\subsection{Adaptive Neuron-wise Discriminant Criterion}

In the neuron-wise discriminant criterion, we have to keep all the features to calculate the within-class variance shown in  Eq. (\ref{sigmaw}) and the between-class variance shown in Eq. (\ref{sigmat}).
In the proposed method, they are calculated by using online computation with the exponential forgetting weights.
The exponential weight is defined as
\begin{align}
    \label{oblivion}
    w(s) = \alpha^{s} \ \ \ \ \  (0 < \alpha < 1).
\end{align}
The weight $w(s)$ becomes smaller when $s$ becomes the bigger.
Then the total mean of the features $\{z_{ik}|i=1,\ldots,n\}$ of the $k$-th neuron is defined by using the forgetting weights as
\begin{align}
    \label{totalmean}
    \mu_{Tk}^{(n)} &= \frac{1}{\sum_{i=1}^n \alpha^{n-i}} \sum_{i=1}^n \alpha^{n-i} z_{ik} \nonumber \\
    &=\alpha \mu_{Tk}^{(n-1)}+(1-\alpha) z_{nk},
\end{align}
where $n$ and $\alpha \in (0,1)$ are the number of samples, and the hyper parameter to define the forgetting rate, respectively.
The last equation gives the update rule of the online computation.
By this online computation, we can compute the total mean for $n$ samples  $\mu_{Tk}^{(n)}$ from the total mean of $n-1$ samples $\mu_{Tk}^{(n-1)}$  and the $n$-th sample $z_{nk}$.

Similar with the total mean shown in Eq. (\ref{totalmean}), we can derive the update rules of online computation for the mean of the target class and the non-target classes of the inputs of $k$-th neuron as
\begin{align}
    \label{targetmean}
    \mu_k^{(n)}&=\frac{1}{\sum_{i=1}^n t_{ik}\alpha^{n-i}} \sum_{i=1}^n t_{ik}\alpha^{n-i} z_{ik} \notag \\
    &=\alpha\mu_k^{(n-1)}+(1-\alpha)t_{nk}z_{nk}, \\
    \label{nontargetmean}
    \hat{\mu}_k^{(n)}&=\frac{1}{\sum_{i=1}^n (1-t_{ik})\alpha^{n-i}} \sum_{i=1}^n (1-t_{ik})\alpha^{n-i} z_{ik} \notag \\
    &=\alpha\hat{\mu}_k^{(n-1)}+(1-\alpha)(1-t_{nk})z_{nk}.
\end{align}
The details of the derivations are shown in Appendix.

Similarly we can derive the update rules of online computation for the within-class and the total variances with forgetting weights as
\begin{align}
    \label{adsigmaw}
    \sigma_{W_k}^{(n)2} &= \frac{1}{\sum_{i}^{n}\alpha^{n-i}}\sum_{i}^{n}\alpha^{n-i}\{t_{ik}(z_{ik}-\mu_k^{(n)})^2\notag\\
    &\ \ \ + (1-t_{ik})(z_{ik}-\hat{\mu}_k^{(n)})^2\}\notag\\
    &= \alpha\sigma_{W_k}^{(n-1)2}+\alpha(1-\alpha)\{t_{nk}(z_{nk}-\mu_k^{(n-1)})^2\notag\\
    &\ \ \ + (1-t_{nk})(z_{nk}-\hat{\mu}_k^{(n-1)})^2\},\\
    \label{adsigmat}
    \sigma_{T_k}^{(n)2} &= \frac{1}{\sum_{i}^{n}\alpha^{n-i}}\sum_{i}^{n}\alpha^{n-i}(z_{ik}-\mu_{Tk}^{(n)})^2\notag\\
    &= \alpha\sigma_{T_k}^{(n-1)2}+\alpha(1-\alpha)(z_{nk}-\mu_{Tk}^{(n-1)})^2.
\end{align}
The details of the derivations are also shown in Appendix.

Then the adaptive neuron-wise discriminant criterion is defined by using Eq.(\ref{adsigmaw}) and Eq.(\ref{adsigmat}) as
\begin{align}
    \label{adaptivediscriminantcriterion}
    \mathcal{L}_{AD} = \sum_{k=1}^{K}\frac{\sigma_{W_k}^{(N)2}}{\sigma_{T_k}^{(N)2}}.
\end{align}

\subsection{Adaptive Center Loss at Hidden Layer}

The original center loss was introduced at the output layer to make the feature more discriminative \cite{t1}, but it is possible to apply it at a hidden layer.
We can observe that the feature vectors extracted before the ReLU function at a hidden layer give the elongated shapes for each class, as shown in Fig. \ref{plot} (a).
In this paper, we introduce the center loss at the hidden layer instead of the output layer to make the features of each class more circular.
Since the neuron-wise discriminant criterion makes the features at the output layer more discriminative, it expected that the recognition accuracy could be improved by combining the center loss at the hidden layer with the neuron-wise discriminant criterion at the output layer.

Also, we propose to use the online computation of the mean vectors of each class with exponential forgetting weights instead of the update computation expressed as Eq.(\ref{pc1}) within mini-batch which is proposed in the original center loss \cite{t1}.

Let $\{\bm{x}_{i} \in \mathbb{R}^D|i=1,\ldots,n\}$ be a set of feature vectors  extracted before ReLU function at a hidden layer shown in Fig.\ref{CNN}, where $D$ is the dimension of an extracted feature vectors. 
Then the total mean vector of these vectors is defined by using the exponential forgetting weights as
\begin{align}
    \label{ac1}
    {\bf c}_k^{(n)} &= \frac{1}{\sum_{i}^{n}\alpha^{n-i}}\sum_{i}^{n}\bm{x}_{i}\alpha^{n-i} \notag \\
    &= \alpha \bm{c}_k^{(n-1)} + (1 - \alpha)\bm{x}_n.
\end{align}
The details of the derivation are shown in Appendix.

The adaptive center loss is defined by using Eq. (\ref{ac1}) as
\begin{align}
    \label{adaptivecenterloss}
    \mathcal{L}_{AC} &= \frac{1}{N}\sum_{i=1}^{N}||\bm{x}_i - \bm{c}_{k}^{(N)}||_2^2.
\end{align}

\section{Experiments}

\begin{table}
    \begin{center}
        \caption{The network architecture used in the preliminary experiments for MNIST dataset.}
          \begin{tabular}{c||c|c|c} \hline
             Layer & Operator & Resolution & Channels \\ \hline \hline
             0 & - & 28x28 & 1 \\ \hline
             1 & Conv3x3+1 & 28x28 & 32 \\ \hline
             1 & RelU & 28x28 & 32 \\ \hline
             1 & Maxpool2x2 & 14x14 & 32 \\ \hline
             2 & Conv3x3+1 & 14x14 & 64 \\ \hline
             2 & RelU & 14x14 & 64 \\ \hline
             2 & Maxpool2x2 & 7x7 & 64 \\ \hline
             3 & FC & 256 & - \\ \hline
             3 & ReLU &256 & - \\ \hline
             4 & FC & 100 & - \\ \hline
             4 & ReLU &100 & - \\ \hline
             5 & FC & 10 & - \\ \hline
          \end{tabular}
          \label{network1}
    \end{center}
\end{table}

\subsection{Preliminary Experiments using MNIST dataset}

To confirm the effectiveness of the proposed approach, we have performed experiments using MNIST dataset.
The activation function in this network is ReLU function in the hidden layers, and the softmax function is used in the output layer.
Two fully-connected layers are used for classification.
The details of the network architecture are shown in  Table \ref{network1}.
In this table, Conv3x3+1 denotes a convolution layer with $3\times 3$ convolution filters, and the stride of the convolution computation is set to 1, and the padding size is 1.
Maxpool2x2 denotes a max-pooling with the size $2\times 2$, and 
FC denotes a fully-connected layer.

The parameters of this network are trained by minimizing the proposed objective function shown in Eq. (\ref{adcenaddisc}).
Stochastic Gradient Descent (SGD) is used with a momentum of $0.9$ as the optimizer. 
Both the number of epochs and the number of samples in mini-batch are set to $100$.
The initial learning rate is set to $0.01$ and is divided by $10$ at every $50$ epochs.
The weight decay parameter is set to $0.01$ to prevent overfitting.

We have performed preliminary experiments to investigate the effectiveness of the proposed approach.
We compared the adaptive neuron-wise discriminant criterion in Eq. (\ref{adaptivediscriminantcriterion})  with the original neuron-wise discriminant criterion \cite{t12} in Eq. (\ref{discriminantcriterion}) and the standard CNN without acceleration for feature discrimination (baseline CNN).
Also, the adaptive center loss in Eq. (\ref{adaptivecenterloss}) at the hidden layer is compared with the original center loss \cite{t1} in Eq.(\ref{centerloss})  at the hidden layer and the standard CNN without acceleration for feature discrimination.

For the neuron-wise discriminant criterion, 10-dimensional features of the output layer are used, and 100-dimensional features before the ReLU activation function at the last fully-connected layer are used to compute the center loss. 
For the neuron-wise discriminant criterion, we set $\lambda$ to $0.01$.
For the center loss, we set $\beta$ and $\lambda$ to $1.0$ and $1.0$, respectively.
For the adaptive neuron-wise discriminant criterion, we set $\alpha$ and $\lambda$ to $0.99$ and $0.01$, respectively.
For the adaptive center loss, we set $\alpha$ and $\lambda$ to $0.99$ and $1.0$, respectively.

\begin{table}
    \begin{center}
        \caption{The results of the preliminary experiment.
        }
          \begin{tabular}{c||c|c|c|c} \hline
                      
                                   & train loss & test loss & \shortstack{train\\accuracy} & \shortstack{test\\accuracy} \\ \hline \hline
            baseline                & 0.0751     & 0.0720    & 0.9808         & 0.9808 \\ \hline
            discriminant           & 0.0849     & 0.0826    & 0.9825         & 0.9817 \\ \hline
            \shortstack{adaptive\\discriminant} & 0.1185     & 0.1216    & 0.9905         & 0.9878 \\ \hline
            center                 & 0.2082     & 0.2605    & 0.9972         & 0.9930 \\ \hline
            \shortstack{adaptive\\center}       & 0.2021     & 0.2489    & 0.9971         & 0.9935 \\ \hline
            \shortstack{adaptive\\discriminant+center}   
                                   & 0.2012     & 0.2477    & \bf{0.9972}    & \bf{0.9937} \\ \hline
          \end{tabular}
          \label{score1}
    \end{center}
\end{table}

Table \ref{score1} shows the results of these experiments.
It is noticed that the adaptive neuron-wise discriminant criterion gives better accuracy than the original neuron-wise discriminant criterion and the baseline CNN, and the adaptive center loss also at a hidden layer gives better test accuracy than the center loss and the baseline CNN.
These results show that the adaptive neuron-wise discriminant criterion and the adaptive center loss at the hidden layer can improve the recognition accuracy.

The recognition accuracy obtained by the integration of the adaptive neuron-wise discriminant criterion and the adaptive center loss at the hidden layer is also included in Table \ref{score1}.
The parameters $\lambda_1$, $\lambda_2$, and $\alpha$ are set to $0.001$, $1.0$, and $0.99$, respectively.
From this table, we can confirm that the proposed integration of the adaptive neuron-wise discriminant criterion and the adaptive center loss at the hidden layer can give the best test accuracy.

\begin{figure}[tb]
\centering
\includegraphics[width=0.7\linewidth]{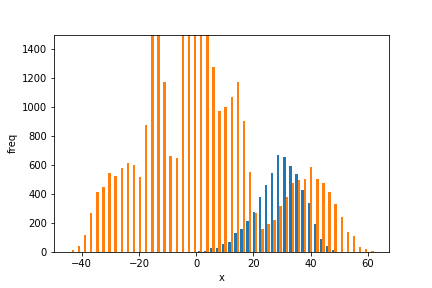}\\
(a) The standard CNN\\
\includegraphics[width=0.7\linewidth]{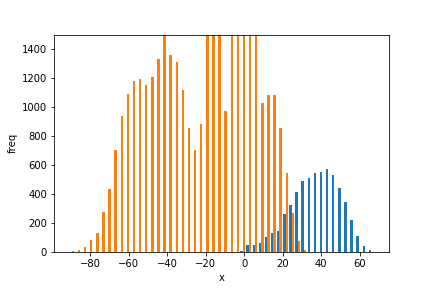}\\
(b) With the adaptive neuron-wise discriminant criterion\\
\includegraphics[width=0.7\linewidth]{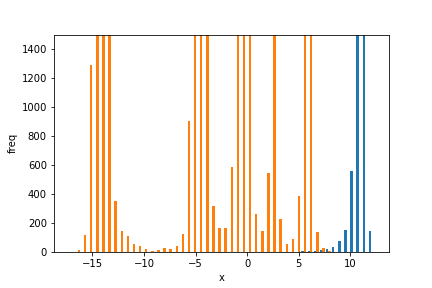}\\
(c) With the adaptive neuron-wise discriminant criterion and the adaptive center loss
\caption{The histograms of the inputs of a neuron at the output layer of the trained CNN. 
Blue shows the histogram of the target class and orange shows the histogram of the non-target class.}
\label{histogram}
\end{figure}

The effectiveness of the proposed approach is confirmed by drawing histograms of the input features of a neuron in the output layer of the CNN which are trained using a training set (60000) of MNIST. 
In this case, the dimension of the feature vector before the ReLU activation function at the last fully-connected layer is changed from $100$ to $2$ in the network architecture shown in Table \ref{network1}. 
Fig. \ref{histogram} (a), (b), and (c) show the histograms obtained by the standard softmax cross-entropy loss, the adaptive neuron-wise discriminant criterion, and the integration of the adaptive neuron-wise discriminant criterion and the adaptive center loss at the hidden layer, respectively.

From this figure, it is noticed that the separations in the histograms (b) and (c) are better than the baseline (a).
This result shows the effectiveness of the adaptive neuron-wise discriminant criterion.
Also, it is noticed that there are multiple distributions in the non-target class (orange).
We think that this phenomenon is caused by the adaptive center loss at the hidden layer.

\begin{figure}[tb]
\centering
\includegraphics[width=0.6\linewidth]{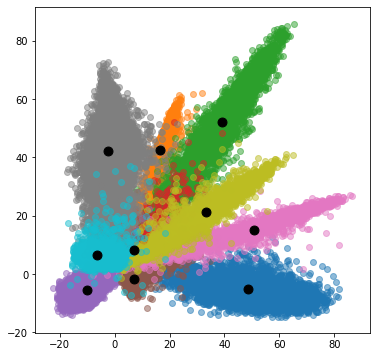}\\
(a) The standard CNN\\
\includegraphics[width=0.6\linewidth]{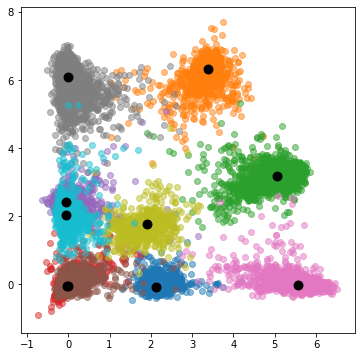}\\
(b) With the adaptive center loss\\
\includegraphics[width=0.6\linewidth]{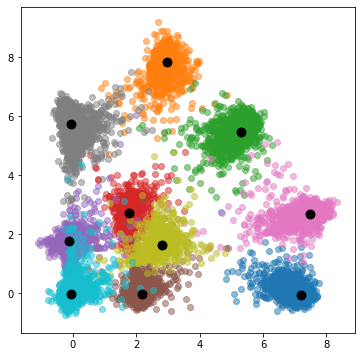}\\
(c) With the adaptive neuron-wise discriminant criterion and the adaptive center loss
\caption{This is the 2-dimensional extracted feature mapping before the ReLU function at the hidden layer learned by CNN with ReLU function using training data (60000) on MNIST. 
Colors indicate the classes and black points are the means of each class.}
\label{plot}
\end{figure}

Fig. \ref{plot} shows the scatter plots of the 2-dimensional feature vectors extracted from the hidden layer learned by CNN with ReLU function using training data (60000) on MNIST.
In this figure, colors indicate the classes, and black points are the means of each class.
Fig. \ref{plot} (a), (b), and (c) are the scatter plots of the feature vectors obtained by using the standard softmax cross-entropy loss, the adaptive center loss, and the integration of the adaptive neuron-wise discriminant criterion and adaptive center loss, respectively.

It is noticed that the distributions of each class are more compact in Fig. \ref{plot} (b) and (c) than the plot shown in Fig. \ref{plot} (a).
This means that the adaptive center loss at the hidden layer is useful to make the feature vectors discriminative.
It is also noticed that the integration of the adaptive neuron-wise discriminant criterion and adaptive center loss can make the feature vectors more discriminative.

\subsection{Comparison Experiments}

\begin{table}
    \begin{center}
        \caption{The network architecture used for comparison experiments.}
          \begin{tabular}{c||c|c|c} \hline
             Layer & Operator & Resolution & Channels \\ \hline \hline
             0 & - & hxw & 3 \\ \hline
             1 & Conv3x3+1 & hxw & 128 \\ \hline
             1 & Batchnorm & hxw & 128 \\ \hline
             1 & RelU & hxw & 128 \\ \hline
             2 & Conv3x3+1 & hxw & 128 \\ \hline
             2 & Batchnorm & hxw & 128 \\ \hline
             2 & RelU & hxw & 128 \\ \hline
             3 & Conv3x3+1 & hxw & 128 \\ \hline
             3 & Batchnorm & hxw & 128 \\ \hline
             3 & RelU & hxw & 128 \\ \hline
             3 & Maxpool2x2 & h/2xw/2 & 128 \\ \hline
             4 & Conv3x3+1 & h/2xw/2 & 256 \\ \hline
             4 & Batchnorm & h/2xw/2 & 256 \\ \hline
             4 & RelU & h/2xw/2 & 256 \\ \hline
             5 & Conv3x3+1 & h/2xw/2 & 256 \\ \hline
             5 & Batchnorm & h/2xw/2 & 256 \\ \hline
             5 & RelU & h/2xw/2 & 256 \\ \hline
             6 & Conv3x3+1 & h/2xw/2 & 256 \\ \hline
             6 & Batchnorm & h/2xw/2 & 256 \\ \hline
             6 & RelU & h/2xw/2 & 256 \\ \hline
             6 & Maxpool2x2 & h/4xw/4 & 256 \\ \hline
             7 & Conv3x3 & (h/4-2)x(w/4-2) & 512 \\ \hline
             7 & Batchnorm & (h/4-2)x(w/4-2) & 512 \\ \hline
             7 & RelU & (h/4-2)x(w/4-2) & 512 \\ \hline
             8 & Conv3x3 & (h/4-4)x(w/4-4) & 256 \\ \hline
             8 & Batchnorm & (h/4-4)x(w/4-4) & 256 \\ \hline
             8 & RelU & (h/4-4)x(w/4-4) & 256 \\ \hline
             9 & Conv3x3 & (h/4-6)x(w/4-6) & 128 \\ \hline
             9 & Batchnorm & (h/4-6)x(w/4-6) & 128 \\ \hline
             9 & RelU & (h/4-6)x(w/4-6) & 128 \\ \hline
             9 & Maxpool2x2 & (h/4-6)/2x(w/4-6)/2 & 128 \\ \hline
             10 & FC & (h/4-6)/2x(w/4-6)/2x128 & - \\ \hline
             10 & ReLU & (h/4-6)/2x(w/4-6)/2x128 & - \\ \hline
             10 & Dropout & (h/4-6)/2x(w/4-6)/2x128 & - \\ \hline
             11 & FC & 100 & - \\ \hline
             11 & RelU & 100 & - \\ \hline
             11 & Dropout & 100 & - \\ \hline
             12 & FC & class & - \\ \hline
          \end{tabular}
          \label{network2}
    \end{center}
\end{table}

The proposed integration of the adaptive neuron-wise discriminant criterion and adaptive center loss is compared with the original neuron-wise discriminant criterion and the center loss at the hidden layer with several datasets such as FashionMNSIT, CIFAR10, CIFAR100, and STL10.
For the FashionMNIST data set, the same network with the preliminary experiment shown in Table \ref{network1}.
The more complex network architecture shown in Table \ref{network2} is used for the data sets CIFAR10, CIFAR100, and STL10.

In Table \ref{network2}, $h$ and $w$ denote the height and the width of input images.
Conv3x3+1 denotes the convolution layer with the size $3\times 3$, and the stride and the padding are 1.
Batchnorm denots batch normalize \cite{t18}.
Maxpool2x2 denotes max-pooling with the size $2\times 2$.
Conv3x3 denotes the convolution layer with the size $3\times 3$, where the stride is 1, but the padding is 0.
FC denotes the fully-connected layer, and ReLU is the activation function.
Dropout denotes drop out \cite{t19}.

These networks are trained by using the training samples of each data set.
For the training, the number of epochs and the number of samples in the mini-batch are set to $500$ and $100$, respectively.
As the optimizer, we use Stochastic Gradient Decent (SGD) with a momentum of $0.9$.
The learning rate is set to $0.01$ and is divided by $10$ at every $100$ epochs.
The weight decay parameter is set to $0.01$ for FashionMNIST, $0.01$ for CIFAR10, $0.001$ for CIFAR100, and $0.01$ for STL10.
As the preprocessing, we apply the affine transformation to the inputs for CIFAR10, CIFAR100, and STL10 to prevent the overfitting.

Similar to the preliminary experiments, each feature of the output layer is used to compute the neuron-wise discriminant criterion, and the 100-dimensional feature vectors before the ReLU function at the hidden layer are used to compute the center loss.

For the neuron-wise discriminant criterion, the hyper parameter  $\lambda$ is set to $0.001$ for FashionMNIST, $0.001$ for CIfAR10, $0.01$ for CIFAR100, and $0.01$ for STL10.
For the center loss, the parameter $\beta$ is set to $1.0$, and the parameter $\lambda$ is set to $1.0$ for FashionMNIST, $0.08$ for CIFAR10, $0.01$ for CIFAR100, and $0.01$ for STL10, respectively.
For the adaptive neuron-wise discriminant criterion and the adaptive center loss, the parameter $\alpha$ is set to $0.99$, and $\lambda_1$ and $\lambda_2$ to $0.0001$ and $1.0$ for FashionMNIST,  $0.0001$ and $0.08$ for CIFAR10, $0.01$ and $0.001$ for CIFAR100, and $0.01$ and $0.001$ for STL10.

The results are shown in Table \ref{score2}.
It is obvious from this Table that the proposed integration of the adaptive neuron-wise discriminant criterion and adaptive center loss at the hidden layer give the best test accuracy for all data sets.
Thus we can say that the proposed approach can improve the classification accuracy of the trained CNN.

\begin{table}
    \begin{center}
        \caption{The results of comparison experiments.
        }
          \begin{tabular}{l||c|c|c|c} \hline
                      
                        & \multicolumn{2}{|c|}{FashionMNIST} & \multicolumn{2}{c}{CIFAR10} \\ \cline{2-5}
                        & train & test & train & test \\ \hline \hline
            baseline     & 0.9223 & 0.9028 & 1.0 & 0.8975 \\ \hline
            discriminant& 0.9238 & 0.9041 & 1.0 & 0.9008 \\ \hline
            center      & 0.9547 & 0.9244 & 1.0 & 0.9089 \\ \hline
            \shortstack{adaptive\\discriminant+center}
                        & 0.9634 & \bf{0.9273} & 1.0 & \bf{0.9148} \\ \hline
             
          \end{tabular}\vspace{3mm}
          \begin{tabular}{l||c|c|c|c} \hline
                      
                        & \multicolumn{2}{|c|}{CIFAR100} & \multicolumn{2}{c}{STL10} \\ \cline{2-5}
                        & train & test & train & test \\ \hline \hline
            baseline     & 0.9334 & 0.6136 & 0.9812 & 0.7463 \\ \hline
            discriminant& 0.9347 & 0.6178 & 0.9986 & 0.7723 \\ \hline
            center      & 0.9387 & 0.6162 & 0.9998 & 0.7690 \\ \hline
            \shortstack{adaptive\\discriminant+center}
                        & 0.9407 & \bf{0.6208} & 0.9986 & \bf{0.7740} \\ \hline
             
          \end{tabular}
          \label{score2}
    \end{center}
\end{table}

\section{Conclusions}
In this paper, we propose to use  the neuron-wise discriminant criterion at the output layer and the center loss at the hidden layer.
Also, we introduce the online computation of the means of each class with the exponential forgetting.
We named them the adaptive neuron-wise discriminant criterion and the adaptive center loss, respectively.
According to Fig.\ref{histogram}, the histogram of each class is more separated by using both.
According to Fig.\ref{plot}, the 2-dimensional extracted feature mapping is more sepatrated by using both.
Through the experiments, we got the effectiveness of the integration of the adaptive neuron-wise discriminant criterion and the adaptive center loss by using the MNIST, FashionMNIST, CIFAR10, CIFAR100, and STL10.

\section*{Acknowledgment}

This work was partly supported by JSPS KAKENHI Grant Number 16K00239.

\appendix

In this section, we show some formulations proof to prove our proposal Eq. (\ref{ac1}), Eq. (\ref{adsigmaw}) and Eq. (\ref{adsigmat}).
We show the proof of updating formulation for general weighted mean and weighted variance.
Let $\{ r_i|i=1\ldots N\}$, and $alpha$ be a set of data, and the parameter to define the forgetting power restricted in $(0,1)$, respectively.
And we suppose that we have infinite 0 data before coming first data.
Then, There are infinite data and weights from $\alpha^{0}$ to $\alpha^{\infty}$.
The summation of geometric sequence is defined as
\begin{align}
    \label{gs}
    S &= \lim_{N\rightarrow\infty}\sum_{i}^{N}\alpha^{N-i} \notag \\
    &=\lim_{N\rightarrow\infty}\frac{1-\alpha^{N+1}}{1-\alpha}\notag \\
    &=\frac{1}{1-\alpha},
\end{align}
where $\lim_{N\rightarrow\infty} \alpha^{N+1} = 0$ if $0 < \alpha < 1$.

By using Eq. (\ref{gs}), the fundamental weighted mean and the fundamental weighted variance are defined as
\begin{align}
    \label{wmean}
    E[r]^{(N)} &= \frac{1}{\sum_{i}^{N}\alpha^{N-i}}\sum_{i}^{N}r_{i}\alpha^{N-i}\notag\\
    &= (1-\alpha)\sum_{i}^{N}r_{i}\alpha^{N-i}\notag\\
    &= (1-\alpha)\alpha \sum_{i}^{N-1}r_{i}\alpha^{N-1-i}+(1-\alpha)r_{N} \notag \\
    &= \alpha E[r]^{(N-1)} + (1 - \alpha)r_N,
\end{align}
\begin{align}
    \label{wvariance}
    V[r]^{(N)} &= \frac{1}{\sum_{i}^{N}\alpha^{N-i}}\sum_{i}^{N}\alpha^{N-i}(r_{i}-E[r]^{(N)})^2 \notag\\
    &= E[r^2]^{(N)} - E[r]^{(N)2} \notag\\
    &= (1-\alpha)\sum_{i}^{N}r_{i}^2\alpha^{N-i} - E[r]^{(N)2} \notag \\
    &= (1-\alpha)\alpha \sum_{i}^{N-1}r_{i}^2\alpha^{N-1-i} + (1-\alpha)r_N^2 - E[r]^{(N)2} \notag \\
    &= (1-\alpha)\alpha \sum_{i}^{N-1}r_{i}^2\alpha^{N-1-i} + (1-\alpha)r_N^2\notag \\
    &\ - (\alpha E[r]^{(N-1)}-(1-\alpha)r_N)^2 \notag \\
    \notag\\
    &= \alpha V[r]^{(N-1)} + \alpha E[r]^{(N-1)2} + (1-\alpha)r_N^2 \notag \\ 
    &\ \ \ - \alpha^2E[r]^{(N-1)2} - 2\alpha (1-\alpha)r_{N}E[r]^{(N-1)}\notag\\
    &\ \ \ - (1-\alpha)^2r_{N}^2\notag\\
    \notag\\
    &= \alpha V[r]^{(N-1)} + \alpha (1-\alpha)(r_{N} - E[r]^{(N-1)})^2,
\end{align}
where $E[r]^{(N)}$ denotes the weighted mean of $N$ times, and $V[r]^{(N)}$ denotes the weighted variance of $N$ times.

The derivation of the update rule Eq.(\ref{totalmean}), Eq.(\ref{targetmean}), Eq.(\ref{nontargetmean}), and Eq.(\ref{ac1}) are leaded by using Eq.(\ref{wmean}).
The derivation of the update rule Eq.(\ref{adsigmaw}) and Eq.(\ref{adsigmat}) are leaded by using Eq.(\ref{wvariance}).


\begin{thebibliography}{40}

\bibitem{t2} He, Kaiming, et al. "Deep residual learning for image recognition." Proceedings of the IEEE conference on computer vision and pattern recognition. 2016.

\bibitem{t3} He, Kaiming, et al. "Delving deep into rectifiers: Surpassing human-level performance on imagenet classification." Proceedings of the IEEE international conference on computer vision. 2015.

\bibitem{t4} Krizhevsky, Alex, Ilya Sutskever, and Geoffrey E. Hinton. "Imagenet classification with deep convolutional neural networks." Advances in neural information processing systems. 2012.

\bibitem{t5} Simonyan, Karen, and Andrew Zisserman. "Very deep convolutional networks for large-scale image recognition." arXiv preprint arXiv:1409.1556 (2014).

\bibitem{t6} Szegedy, Christian, et al. "Going deeper with convolutions." Proceedings of the IEEE conference on computer vision and pattern recognition. 2015.

\bibitem{t7} Zhou, Bolei, et al. "Object detectors emerge in deep scene CNN." arXiv preprint arXiv:1412.6856 (2014).

\bibitem{t8} Zhou, Bolei, et al. "Learning deep features for scene recognition using places database." Advances in neural information processing systems. 2014.

\bibitem{t9} Baccouche, Moez, et al. "Sequential deep learning for human action recognition." International workshop on human behavior understanding. Springer, Berlin, Heidelberg, 2011.

\bibitem{t10} Ji, Shuiwang, et al. "3D convolutional neural networks for human action recognition." IEEE transactions on pattern analysis and machine intelligence 35.1 (2012): 221-231.

\bibitem{t11} Wang, Limin, Yu Qiao, and Xiaoou Tang. "Action recognition with trajectory-pooled deep-convolutional descriptors." Proceedings of the IEEE conference on computer vision and pattern recognition. 2015.

\bibitem{t12} Ide, Hidenori, and Takio Kurita. "Convolutional Neural Network with neuron-wise discriminant criterion for Input of Each Neuron in Output Layer." International Conference on Neural Information Processing. Springer, Cham, 2018.

\bibitem{t1} Wen, Yandong, et al. "A discriminative feature learning approach for deep face recognition." European conference on computer vision. Springer, Cham, 2016.

\bibitem{t13} Hoffer, Elad, and Nir Ailon. "Deep metric learning using triplet network." International Workshop on Similarity-Based Pattern Recognition. Springer, Cham, 2015.

\bibitem{t15} Wang, Hao, et al. "Cosface: Large margin cosine loss for deep face recognition." Proceedings of the IEEE Conference on Computer Vision and Pattern Recognition. 2018.

\bibitem{t17} Li, Li, Miloš Doroslovački, and Murray H. Loew. "Discriminant Analysis Deep Neural Networks." 2019 53rd Annual Conference on Information Sciences and Systems (CISS). IEEE, 2019.

\bibitem{t16} Nair, Vinod, and Geoffrey E. Hinton. "Rectified linear units improve restricted boltzmann machines." Proceedings of the 27th international conference on machine learning (ICML-10). 2010.

\bibitem{t18} Ioffe, Sergey, and Christian Szegedy. "Batch normalization: Accelerating deep network training by reducing internal covariate shift." arXiv preprint arXiv:1502.03167 (2015).

\bibitem{t19} Srivastava, Nitish, et al. "Dropout: a simple way to prevent neural networks from overfitting." The journal of machine learning research 15.1 (2014): 1929-1958.

\end{thebibliography}
\end{document}